# Ant Detective: An Automated Approach for Counting Ants in Densely Populated Images and Gaining Insight into Ant Foraging Behavior

Mautushi Das[1], Fang-Ling Chloe Liu[2], Charly Hartle[2], Chin-Cheng Scotty Yang[2*], and C. P. James Chen[1*]

[1]School of Animal Sciences, Virginia Tech, VA 24061, USA, [2]Department of Entomology, Virginia Tech, VA 24061, USA

*Corresponding author:

C. P. James Chen <niche@vt.edu> and Chin-Cheng Scotty Yang <scottyyang@vt.edu>

Acknowledgment

*This project received partial funding from an internal competitive grant from the College of Agriculture and Life Sciences at Virginia Tech. The authors have declared no conflict of interest. We acknowledge the use of ChatGPT-4 in organizing and articulating ideas during the preparation of this manuscript. While the intellectual contributions and conceptual developments are entirely those of the authors, we duly recognize ChatGPT's role in streamlining the writing process.*

Abstract

Ant foraging behavior is critical for understanding ecological dynamics and developing effective pest management strategies. However, quantifying this behavior is challenging due to the labor-intensive nature of manual counting, particularly in densely populated images. This study introduces an automated approach using computer vision to count ants and analyze their foraging patterns. Leveraging the YOLOv8 model, the system was calibrated and tested on datasets with varying imaging scenarios and densities. The results show that the system achieves an average precision and recall of up to 87.96% and 87.78%, respectively, with just 64 calibration images, provided both calibration and evaluation images share similar backgrounds. In cases where the evaluation images have more complex backgrounds than the calibration set, the system requires a larger calibration dataset to generalize effectively. Using 1,024 calibration images in such cases, the system achieved a precision of 83.60% and a recall of 78.88%. For particularly challenging scenarios where over a thousand ants appear in a single image, the system can still maintain satisfactory performance by slicing images into smaller patches, reaching a precision of 77.97% and a recall of 71.36%. Additionally, the system generates heatmaps that visualize the spatial distribution of ant activity over time, offering valuable insights into their foraging behavior. This spatial-temporal analysis deepens our understanding of ant behavior, aiding both ecological research and pest control efforts. By automating the counting process and providing detailed behavioral insights, this study offers an efficient tool for researchers and pest control professionals to develop more effective strategies.

Introduction

Ant foraging is probably one of the most studied themes in ant research because their behavioral repertoires are remarkably diverse (Reeves and Moreau 2019). Studying foraging behavior in ants can facilitate our understanding of ecological dynamics, as ants play vital roles in numerous ecosystems functioning including nutrient cycling, seed dispersal, etc. (Parr and Bishop 2022). Their foraging patterns can also reveal insights into social organization, communication, and adaptability to environmental changes including pathogen challenge (Alciatore et al. 2021) or pesticide exposure (Thiel and Köhler 2016). Foraging behaviors are particularly crucial when developing an effective baiting strategy. This is because the success of baiting heavily relies upon the ants ingesting a lethal dose of the bait, which can be readily facilitated by taking advantage of better understanding of their foraging behavior (Galante et al. 2024).

Quantifying long-term behaviors and investigating foraging preferences in ants typically involves manual counting across a large number of images. This process is generally labor-intensive, time-consuming and prone to human error, particularly when dealing with large numbers of ants. As a result, scaling up experiments to simulate more natural conditions, such as larger colony sizes and increased numbers of colonies, is challenging. For example, Hsu et al (2018) observed foraging behavior on a total of 18 red imported fire ant (*Solenopsis invicta*) colonies, with each colony comprising approximately 3,000 individual ants. Each colony was offered four food resources with different macronutrient ratios. The number of foragers on each food source was manually estimated by analyzing photos taken by an automatic recording system that takes a picture on each food resource every 10 minutes for 32 hours, which resulted into a total of 13,824 photos that were processed by the study. Additionally, the number of ants counted in an image or at a specific time point may not fully capture the complexity of foraging behaviors, which often involve both spatial factors and the interaction between spatial and temporal dynamics, such as testing different sugar concentrations (Sola and Josens 2016) or bait types (Du et al. 2023).

These challenges make manual counting a laborious task. While it can still yield results, tools that specifically tackle this challenge would facilitate counting process and thus the development of more effective pest control strategies. Indeed, relevant efforts have been made to automate the process of counting insects. Thresholding is a common method used to segment objects in images, enabling automated insect counting. It assumes that the object of interest exhibits distinct visual contrast with the background. For example, black fly adults were automatically counted by being placed on a Petri dish with a white background after collection from light traps (Parker et al. 2020). Similarly, Smythe et al. (2020) used a comparable method to count horn flies on cattle in the field, relying on the color contrast between the flies and the cattle's skin.

However, this method can be biased when the object and the background share similar visual properties. A more advanced approach involves using deep learning models that learn the morphological patterns of the object and are less sensitive to background appearance. For instance, the model YOLOv7 (Wang et al. 2022) and YOLOv8 (Jocher et al. 2023; Terven et al. 2023) were leveraged to count polyphagous moths in a customized trap (Saradopoulos et al. 2023). However, it can still be challenging to detect small objects like ants due to the limitations of current computer vision (CV) models, which are the primary approach for extracting information from images or videos. The challenges can be attributed to two main factors: pretraining data sources and model architecture. Typically, the pretraining data sources used to establish and train CV models are public datasets like COCO (Lin et al. 2014) and ImageNet (Deng et al. 2009), designed for larger objects occupying a significant portion of an image. Since ants are small and often densely packed, standard models struggle to identify them accurately. The model architecture is another factor that limits the detection of small objects. Most CV models, such as YOLOv8, first resize input images to 640 × 640 pixels and then process them through several convolutional layers, resulting in progressively smaller feature maps. The smallest of these feature maps has a spatial dimension of 20 × 20 pixels. This downsampling means that an object smaller than 32 × 32 pixels (calculated by dividing 640 by 20) will be reduced to a single value in the smallest feature map. As a result, the model cannot effectively capture the fine spatial details necessary to distinguish small objects like ants, making accurate detection extremely difficult.

To address the limitation, a study introduced Fully Convolutional Regression Networks (FCRNs) to count small insects, such as thrips and aphids, on leaves (Bereciartua-Pérez et al. 2023). The FCRN architecture is a symmetric encoder-decoder network that overcomes the size limitation of feature maps by upsampling them to the original image size, preserving the spatial information necessary for accurate counting. Additionally, the output of the FCRN is a density map that shares the same spatial dimensions as the input image and contains only the objects' centroids. Instead of predicting the exact size and location of each object, as in most standard object detection models, the FCRN estimates the probability of an object's presence at each pixel location. The disadvantage of this approach is that it usually requires a large amount of training data to generalize well to new images due to the lack of pre-trained models on public datasets containing millions of images. Another strategy to improve the detection of small objects is slicing the input image into smaller patches and then feeding them separately into the model. This approach also overcomes the feature map size limitation, as it resizes the smaller patches to a larger size before processing (Hong et al. 2021; Bereciartua-Pérez et al. 2023).

Considering these challenges and the potential benefits of automated counting systems, this study explores the use of CV to automate the counting process and provide a more comprehensive analysis of both the spatial and temporal dimensions of ant foraging behavior. The study has three main objectives: (1) determining the quantity of image data required for the system to generalize to new images under similar or different imaging conditions, (2) examining the system's performance in densely packed imaging scenarios, and (3) investigating how CV can enhance our understanding of the spatial and temporal aspects of ant foraging behaviors.

## Materials and Methods

### *Experimental Setup*

*Tapinoma sessile* workers were allowed to feed on sucrose and/or peptone solutions using different setups (**Figure 1**). A01 consisted of a capped Petri dish containing 5 mL of 10% sucrose solution. A cotton thread, with one end submerged in the sucrose solution and the other end resting on the Petri dish lid, allowed the ants to feed while minimizing evaporation. In A02, worker ants were provided with two liquid food dispensing devices identical to those in A01. One contained 10% sucrose solution, while the other contained 5% peptone solution. A03 consisted of a Petri dish with an upside-down 5 mL vial filled with 10% sucrose solution. Three cotton threads extended from the vial to the Petri dish lid, allowing the ants to access the sucrose. In all A01–A03 setups, the feeding devices were placed in a simple foraging area (a clean 15-cm cylindrical container) connected to the ant colony via a plastic tube. We then set up additional experiments with more complex background: B01 was similar to A03, but used a triangular-shaped filter paper instead of cotton threads to dispense the sucrose solution. The device was placed in a foraging arena scattered with debris and parts of insect prey to add complexity to the background; In B02, both virus-infected (odorous house ant virus 1, OHAV-1) and uninfected *T. sessile* ants were allowed to forage from a liquid food dispensing device similar to A02 (offering 10% sucrose and 5% peptone solutions), but with a more complex background. In B03, virus-infected (Solenopsis invicta virus 3, SINV-3) and uninfected lab cultures of the red imported fire ant (*Solenopsis invicta*) were used to generate images. A lab fire ant culture is housed in a tray that has a 15-cm Petri dish painted in red as the nest cell and two tubes with water trapped inside as water feeders. To capture the number of workers that stay outside of the at different timepoints, a photo of entire lab culture was taken at 5 pm each day using the automated recording system (see below for detailed system setup).

*Data Collection and Image Annotation*

The image dataset used in this study was divided into seven distinct subsets, as illustrated in **Figure 1a** and **Table 1**. **Table 1** summarizes the properties of each subset, including the number of images and the average number of ants per image. The dataset was captured using an automated recording system comprises a GoPro camera mounted with a 15X macro lens, which was clamped on the desk edge via a gooseneck mount, allowing the camera to be approximately 10 cm on the top of the ant's foraging arena where foragers were feeding. All images were taken under consistent lighting conditions to ensure uniform image quality and maintained a minimum resolution of 1920 x 1080 pixels.

The "Calibration" subset shares a similar imaging background with the subsets labeled "A" (i.e., "A01", "A02", and "A03"). In contrast, the "B" subsets exhibit more complex imaging conditions. "B01" includes images with debris and parts of insect prey resembling ants, "B02" contains images with non-uniform backgrounds, and "B03" represents images with dense ant populations containing more than 700 ants per image on average.

The manual counting was achieved by annotating the images using the YOLO object detection format (Jocher et al. 2023), which is also a data format for calibrating the CV system to recognize ants. As depicted in **Figure 1b**, in this format, each ant is marked with a bounding box, defined by four parameters: the x and y coordinates of the box's center, along with its width and height. These values are normalized to the range [0, 1] by dividing the x and y coordinates by the image's width and height, respectively. For instance, in a 1920 x 1080 image, a bounding box with a center at (960, 540) and dimensions of 100 x 100 pixels would be represented as (0.5, 0.5, 0.0521, 0.0926), where 0.0521 and 0.0926 are the normalized width and height. Each ant was assigned a class ID. Since the study focuses exclusively on detecting ants without differentiating between species, all ants were assigned a class ID of 0.

*Objective 1: Determining the Amount of Image Resources Required for Generalization*

The "Calibration" subset was used to "teach" the CV system how the ant morphology appears in images, while the other subsets were employed to evaluate its generalization capabilities. The "A" subsets were designed to mimic the system being deployed in an ongoing and repetitive experiment, with images with similar imaging backgrounds over time. In contrast, the "B" subsets aim to test the system robustness in handling images with less controlled imaging conditions. Except for the subset "B03", in which the number of ants per image was significantly higher than in the other subsets and required additional procedures to calibrate the system, this study focused on the performance of the CV system with different numbers of available images for the model calibration. The available images were randomly sampled from the "Calibration" subset given the numbers of 64, 128, 256, 512, and 1024. Since the number of available images was only 954, when the sampled number was set as 1024, the system will sample the images with replacement until the number of images reaches 1024. The sampling process was repeated 30 times for each subset and each available number to avoid sampling bias.

*Objective 2: Strategies for Handling Dense Imaging Scenarios*

Detecting ants in a dense population, such as in subset "B03", is challenging despite an abundance of image data, due to the limitations of the model architecture described in the introduction. Inspired by the Slicing Aided Hyper Inference technique (Akyon et al. 2022), which improves detection accuracy by dividing input images into smaller patches for separate object detection, this study explores an optimized patch size for subset "B03". To avoid bias from varied sampling, all available images—excluding the "B03" subset—were used for model calibration. The system was calibrated using 1,597 images containing 16,368 total ant instances.

During the calibration process, multiple candidate model weights were generated, and the model with the best performance on two randomly selected images from the "B03" subset was selected for further evaluation. These two images were excluded from subsequent evaluation steps. Finally, the model was evaluated on the "B03" subset with different slicing size. The original images, with a resolution of 1636 x 2180 pixels, were divided into different patch sizes: 818 x 1090 (2 x 2 patches), 818 x 545 (2 x 4 patches), 409 x 545 (4 x 4 patches), 409 x 218 (4 x 10 patches), and 204

x 218 (8 x 10 patches), as shown in **Figure 2**. The evaluation aimed to identify the optimal patch size for this dense imaging scenario.

*Objective 3: Enhancing Understanding of Spatial and Temporal Aspects of Ant-Foraging Behaviors*

In the object detection format, the precise location and size of each ant in an image are tracked, allowing the CV system to provide more detailed insights into ant foraging behavior than manual counting, which only captures the number of ants in an image. This study utilizes this data format and a Gaussian inference approach to generate an ant activity heatmap, visualizing the spatial distribution of ant activity across images taken over hours to days.

The first step involves converting each detection's bounding box into a circle, which contains the $(x_0, y_0)$ coordinates and their width and height. The center of the circle is at $(x_0, y_0)$, and the radius (**r**) is the average width and height. An all-zero grid with a resolution of 1000 x 1000 pixels, which are arbitrarily selected, is then initialized as a placeholder matrix for the heatmap. For each ant detection, the Euclidean distance (**d**) from the center of the circle to each pixel in the grid is calculated (**Eq. 1**):

$$d(x, y) = \sqrt{(x - x_0)^2 + (y - y_0)^2}$$

The squared distance, **d²**, is then divided by the squared radius, **r²**, to determine an inverse intensity value. Greater distances correspond to lower intensities. The inverse intensity values are exponentiated to ensure a smooth gradient representation for each ant (**Eq. 2**). These values $G(x, y)$ are accumulated on the grid for all ant detections across all images, producing a heatmap that visually represents areas of higher ant activity.

$$G(x, y) = e^{-(\frac{d(x,y)}{2r^2})}$$

In addition to spatial data, temporal changes in ant presence were analyzed to examine the relationship between the number of ants and time during the study period. For example, in subset "B02," where two bait types, sucrose (labeled 'S') and peptone (labeled 'P'), were placed in the same image, the aim was to understand the ants' foraging preferences. If the sucrose bait was positioned in the upper-right and the peptone in the lower-left of the image, a simple linear function (Eq. 3) was used to separate the ants based on their attraction to the respective baits:

$$\mathcal{L}(x, y) = y - ax - b$$

Where $a$ is the slope, $b$ is the intercept, and $(x, y)$ represents the center of the ant detection. If the result of L(x, y) is positive, the ant is located on the upper (and right when the slope is positive) side of the line and is attracted to the sucrose; if negative, the ant is on the lower (and left when the slope is positive) side of the line and is attracted to the peptone.

*Model Calibration and Evaluation*

The CV detection system is based on the YOLOv8 architecture (Jocher et al. 2023). Given that ant detection is relatively simple compared to general object detection tasks involving over 50 object classes (Everingham et al. 2010; Lin et al. 2014), the smallest model version, YOLOv8n, with 3.2 million parameters, was chosen to balance detection accuracy and computational efficiency. This model version is suitable for deployment on most personal computers due to its minimal hardware requirements (e.g., GPU or high memory), without compromising detection performance, particularly for simple tasks focusing on one specific object (Das et al. 2024).

The model was calibrated using the designated number of images and subsets from Studies 1 and 2. Data augmentation was applied during the calibration process to mitigate imaging biases caused by inconsistent camera angles, lighting conditions, and ant distribution. This augmentation involved introducing random noise, rotation, scaling, and cropping to the original images, enhancing the model's robustness to such variations. The model was calibrated with the Adam optimizer (Kingma and Ba 2017), using a learning rate scheduler that started at 0.001 and decreased by 10% every 10 epochs. The batch size was set to 16, and the calibration was conducted for a total of 100 epochs. Twenty percent of the calibration dataset was randomly selected as the validation set to monitor performance during calibration. The model achieving the best performance on the validation set was chosen for subsequent evaluation. Calibration was conducted using the Ultralytics framework (Jocher et al. 2023) on NVIDIA A100 GPUs.

Model evaluation was performed on the designated subsets from Studies 1 and 2, using metrics such as recall, precision, correlation $r^2$, and Root Mean Square Error (RMSE). Precision and recall were calculated based on the number of true positive (TP), false positive (FP), and false negative (FN) detections:

$$Precision = \frac{TP}{TP + FP}$$

$$Recall = \frac{TP}{TP + FN}$$

A high recall indicates that the model successfully detected most ants in the images, while a high precision indicates that the detections were mostly correct with few false positives. Two additional criteria were considered when calculating precision and recall: Intersection over Union (IoU) and confidence threshold. IoU measures the overlap ratio between the detected bounding box and the actual area occupied by the ant, while the confidence threshold is the minimum confidence score required for a detection to be included in the results. This study set the IoU and confidence threshold at 0.6 and 0.25, respectively. In addition to evaluating detection performance, $r^2$ and RMSE were calculated to compare the automated counting results with the manual counts when the all 954 available calibration images were used:

$$r^2 = \left(\frac{cov(\hat{y}, y)}{\sigma_{\hat{y}}\, \sigma_y}\right)^2$$

$$RMSE = \sqrt{\frac{\sum_i^n (\hat{y}_i - y_i)^2}{n}}$$

Where y and $\hat{y}$ are the manual and automated counts, respectively, and n is the total number of images. $cov(\hat{y}, y)$ is the covariance between $\hat{y}$ and y, and $\sigma$ represents the standard deviations of the term. The r² assesses the agreement between automated and manual counting results, while RMSE measures their absolute difference. Depending on different needs of model accuracy, such as focusing on precise localization or counting, these metrics provide a comprehensive evaluation of the model's effectiveness and reliability for automated ant detection and counting

## Results and Discussion

*Relationship between Model Performance and Calibration Sample Sizes*

The model performance in terms of precision and recall is presented in **Table 2** and **Figure 3**. With a calibration set consisting of only 64 images, similar in background to the test images, the model demonstrates reasonably good performance on subsets A01, A02, and A03. The average precision achieved is 87.96%, 76.01%, and 76.76%, respectively, while the average recall is 87.78%, 77.58%, and 70.23%, respectively. However, increasing the calibration set size does not consistently improve model performance. For instance, in subset A03, doubling the calibration size from 128 to 256 images results in only a slight improvement of 0.22% in precision and 0.41% in recall. Even when increasing the calibration size twentyfold, from 64 to 1024 images, the maximum gains observed are 7.54% in precision for subset A02 and 11.64% in recall for subset A03. In subset A01, these improvements are even smaller, at just 5.14% and 5.33%, respectively. When detection an image containing 20 ants, the 5% improvement in precision and recall corresponds to correctly detecting just one additional ant. These findings suggest the presence of a saturation point in model performance; when the detection task is relatively straightforward and the background is consistent, a small number of calibration images is sufficient to achieve good performance.

In contrast, for more complex and diverse backgrounds, the relationship between sample size and model performance varies significantly between subsets B01 and B02. In subset B01, the increase in performance is almost linear as the calibration size doubles, with precision and recall improving by 14.64% and 13.09%, respectively. Notably, when all calibration images are utilized, the model's performance on B01, despite the new and complex background, is nearly equivalent to that on subsets A02 and A03, which feature similar backgrounds. This suggests that deploying the CV system in a new environment requires a large calibration set to ensure the model can generalize effectively to novel images. Interestingly, in subset B02, characterized by sparse ant distribution and uneven background colors, increasing the calibration size does not significantly enhance model performance. In fact, when the calibration size exceeds 512 images, precision decreased. This finding indicates that the model may overfit to the calibration set when the background is highly diverse, leading to generating more false detections and reducing precision.

When comparing manual and automated counting results (**Figure 4**) using a calibration size of 1024, all subsets except B02 show a high correlation between the automated and manual counts, with a minimum squared correlation ($r^2$) of 0.94. For subsets with similar backgrounds, the RMSE values are 0.96, 1.55, and 1.23 for A01, A02, and A03, respectively. In subset B01, where the background includes multiple debris and parts of insect prey resembling ants, the automated counts achieve an r² of 0.95 but with a higher RMSE of 11.36, indicating an overestimation by approximately 11 ants on average. This overestimation is primarily due to false detections caused by background objects? that resemble ants. The poorest performance is observed in subset B02, where the RMSE is 1.50 and r² drops to 0.58. This suggests that the sparse and small distribution of ants makes the counting results highly sensitive to false detections, significantly affecting the correlation. These findings highlight the importance of background uniformity and ant distribution in ensuring accurate automated counting results.

*Slicing Dense-populated Images Significantly Improves Model Performance*

The CV system achieved a substantial accuracy improvement by slicing densely populated images into smaller patches (**Figure 5**, **Table 3**). Peak performance was observed when the original images were divided into 4 x 10 patches, resulting in a precision of 77.97% and a recall of 71.36%. This represents a significant enhancement compared to the performance on the original images (precision: 9.92%, recall: 1.60%). Additionally, since the YOLOv8n model architecture requires the input image to be in a 640x640 matrix, the slicing ratio plays a crucial role in preventing image distortion, especially when the height/width ratio of the input image deviates from 1. For instance, in this study, the original image size was 1636x2180 pixels. Directly resizing this image to 640x640 results in a height-axis distortion of 1.34 times (calculated from the width/height ratio of 1636/2180). The study also examined whether such height-axis distortion, ideally close to 1, would affect detection performance. Surprisingly, the 4 x 10 patched images, which exhibited the most distorted ratio of 0.53, achieved the best performance, while the 8 x 10 patched images, which were the least distorted at 1.06 times, performed the worst among the sliced images. This indicates that height-axis distortion has no significant correlation with detection performance. **Figure 6** provides an example of detection results between the original image and the 4 x 10 patches. Most missed detections in the original image occurred where ants were densely populated, such as on the edges of the nest and the water feeder. This example demonstrates the effectiveness of the slicing strategy, as only a few ants were missed in the 4 x 10 patched images. The automated counting results for the 4 x 10 patched images were compared with manual counts, showing a high correlation with an $R^2$ of 0.938 and an RMSE of 465.05 (**Figure 7a**, left). The high RMSE value is attributed to overlooked ants in the manual counts, while the CV system was able to capture these missing ants, maintaining strong agreement with the manual counts. The experiment in B03 was to test whether SINV-3 infection alters foraging behaviors in fire ants. The automated counting results effectively captured the ant foraging dynamics in the presence of SINV-3 infection (**Figure 7a**, right). Moreover, when propagating the automated counts to the entire image set collected over the 14-day experiment, a temporal trend of ant population dynamics was observed, which closely matched the manual counts (**Figure 7b**). Both counting methods revealed a similar trend in ant foraging dynamics, with SINV-3 infected ants consistently showing higher levels of foraging behavior than the uninfected group. This result demonstrates the potential of the CV system to accurately capture ant foraging dynamics in a high-throughput manner, which is crucial for studying the effects of pathogens on foraging patterns and potentially other complex social behavior.

*Combining Spatial and Temporal Information Enhances Ant Detection and Counting*

The study utilized a series of time-ordered images to create a heatmap illustrating the spatial distribution of ants over time (**Figure 8**). In experiment B01, three trials were conducted to assess the impact of liquid food dispensing design on ant foraging behavior (**Figure 8a**). In the first trial, a triangular filter paper was placed to distribute sugar solution from a 5mL vial. In the second and third trials, a thread was used to dispense sucrose solutions from a 15mL Falcon tube. The heatmap for the first trial indicates that ants were evenly distributed around the filter paper, with higher densities observed along the edges between the filter paper and the petri dish. For the thread-guided strategy, both trials showed high activities around the thread and the cap of the tube, with the second trial displaying a more concentrated ant distribution along the thread. Analysis of ant distribution over time revealed higher numbers of foraging ants in the first trial, suggesting that the triangular filter paper, which has a larger total area to dispense liquid food than the threads in other trials, effectively attracted more foraging ants.

Another experiment, B02, was conducted to investigate the macronutrient preference of OHAV-1 infected *T. sessile* (**Figure 8b**). Both uninfected and OHAV-1 infected ants were observed. A hypothetical line was drawn between the two food sources (i.e., Petri dishes) to analyze their preferences. The heatmap showed that infected ants foraged more on sucrose, while uninfected ants exhibited no obvious preference between the two macronutrients. When ant presence was plotted over time, it was observed that infected ants exhibited strong foraging preference and activities during the first five hours, followed by a sharp decline. This result demonstrates the capability of the CV system to capture spatial and temporal information simultaneously, providing insights into ant foraging behaviors that could be challenging to obtain through manual counting methods when ant colonies of large size.

The dense ant images from experiment subset B03 were visualized in a heatmap using detection results from the 10 x 4 patched images (**Figure 9**). This experiment examined the impact of SINV-3 infection on ant foraging behaviors, but with a significantly denser ant population and an ant nest cell positioned at the center of the image. The heatmap does not show a strong difference in foraging intensity between the uninfected and infected ants; both groups display hotspots in the four corners of the container, reflecting their natural tendency to explore the environment. Additionally, the heatmap indicates that workers ants are more concentrated around water sources compared to food sources and the nest area. This information is essential for understanding ant foraging behavior and can be applied to optimize bait placement in pest control strategies.

*Data dissemination and web-based application*

The source code, the dataset organized in YOLO format, along with the calibrated YOLOv8n model weights (.pt), can be accessed for generating and evaluating the CV system is publicly available on GitHub (https://github.com/Niche-Squad/ant-detective). To support researchers and practitioners, a web-based Streamlit application called Ant Detective (https://ant-detective.streamlit.app) has also been developed. This application allows users to upload images and receive a folder containing detection results in YOLO format and visualized images with blue bounding boxes.

## Conclusion

This study successfully demonstrated the use of computer vision, specifically the YOLO model, to automate ant detection and counting, offering a more efficient and comprehensive alternative to manual methods. The findings reveal that the quantity of image data required for model calibration varies depending on the complexity of the background. In scenarios with consistent backgrounds, a relatively small calibration set proves sufficient, while more complex backgrounds necessitate a larger calibration set for effective model generalization. The study also highlighted the effectiveness of image slicing in enhancing model performance when dealing with dense ant populations. By dividing the image into smaller patches, the model can overcome the limitations of standard feature map sizes and achieve higher accuracy in detecting individual ants. The ability to track the precise location and size of each ant using the YOLO object detection format enables the generation of ant activity heatmaps, providing valuable insights into the spatial and temporal dynamics of ant foraging behavior. The integration of spatial and temporal information offers a deeper understanding of foraging patterns, potentially aiding in the development of more targeted and effective pest control strategies.

Tables

*Table 1.* Summary information of each image subset

| Subset names | Number of Images | Average Number (Ants per Image) | Standard Deviation (Ants per Image) | Note |
|---|---|---|---|---|
| Calibration | 954 | 9.20 | 7.37 | Similar to the subsets A1, A2, and A3 |
| A01 | 378 | 4.59 | 3.83 | Single bait source |
| A02 | 289 | 4.38 | 4.30 | Two bait sources |
| A03 | 56 | 6.11 | 6.15 | Tube feeder |
| B01 | 151 | 13.75 | 13.94 | Tube feeder with complex background |
| B02 | 206 | 0.69 | 1.26 | Outdoor and complex background |
| B03 | 60 | 717.70 | 442.76 | Dense-populated ants with their colony |

*Table 2. Model performance on different evaluation subsets with varying calibration set sizes (n). The precision and recall values are presented as the mean ± 1.96 standard deviations from 30 sampling iterations. The Intersection over Union (IoU) threshold and confidence threshold are set to 0.6 and 0.25, respectively. The highest precision and recall values of each subset are highlighted in bold.*

| Subset | n | Precision | Recall |
|---|---|---|---|
| A01 | 64 | 87.96% ± 3.72% | 87.78% ± 4.00% |
|  | 128 | 91.07% ± 1.84% | 90.45% ± 1.94% |
|  | 256 | 92.32% ± 1.71% | 91.06% ± 1.78% |
|  | 512 | 92.80% ± 2.27% | 92.65% ± 1.38% |
|  | 1024 | **93.10% ± 1.60%** | **93.11% ± 1.60%** |
| A02 | 64 | 76.01% ± 6.92% | 77.58% ± 6.82% |
|  | 128 | 80.27% ± 8.17% | 82.20% ± 5.57% |
|  | 256 | 81.60% ± 5.35% | 83.68% ± 4.49% |
|  | 512 | 82.31% ± 4.43% | 85.17% ± 3.63% |
|  | 1024 | **83.55% ± 4.27%** | **85.67% ± 3.96%** |
| A03 | 64 | 76.76% ± 7.72% | 70.23% ± 11.72% |
|  | 128 | 83.52% ± 5.49% | 76.33% ± 6.06% |
|  | 256 | 83.74% ± 5.15% | 76.74% ± 6.92% |
|  | 512 | 86.78% ± 3.61% | 80.40% ± 4.70% |
|  | 1024 | **87.79% ± 4.57%** | **81.87% ± 3.84%** |
| B01 | 64 | 68.96% ± 7.47% | 65.79% ± 7.62% |
|  | 128 | 75.48% ± 7.82% | 70.26% ± 5.72% |
|  | 256 | 77.91% ± 5.96% | 73.68% ± 3.43% |
|  | 512 | 80.75% ± 6.25% | 76.46% ± 4.12% |
|  | 1024 | **83.60% ± 2.49%** | **78.88% ± 3.00%** |
| B02 | 64 | 75.04% ± 30.97% | 52.75% ± 15.33% |
|  | 128 | 76.89% ± 24.87% | 57.22% ± 12.13% |
|  | 256 | **78.54% ± 13.47%** | 59.02% ± 10.86% |
|  | 512 | 76.30% ± 18.80% | 59.35% ± 8.08% |
|  | 1024 | 75.16% ± 23.25% | **60.58% ± 11.09%** |

*Table 3. Model performance on original and sliced images with varying patch ratios, resolutions, and height-to-width ratios. Precision and recall values are calculated using an Intersection over Union (IoU) threshold of 0.6 and a confidence threshold of 0.25. The highest precision and recall values are highlighted in bold.*

| Patch | Resolution (pixels) | Height-to-width ratio | Precision | Recall |
|---|---|---|---|---|
| Original | 1636 x 2180 | 1.34 | 9.92% | 1.60% |
| 2 x 2 | 818 x 1090 | 1.34 | 62.15% | 47.73% |
| 2 x 4 | 818 x 545 | 0.67 | 72.33% | 60.88% |
| 4 x 4 | 409 x 545 | 1.34 | 75.26% | 65.35% |
| 4 x 10 | 409 x 218 | 0.53 | **77.97%** | **71.36%** |
| 8 x 10 | 205 x 218 | 1.06 | 41.75% | 37.31% |

Figures

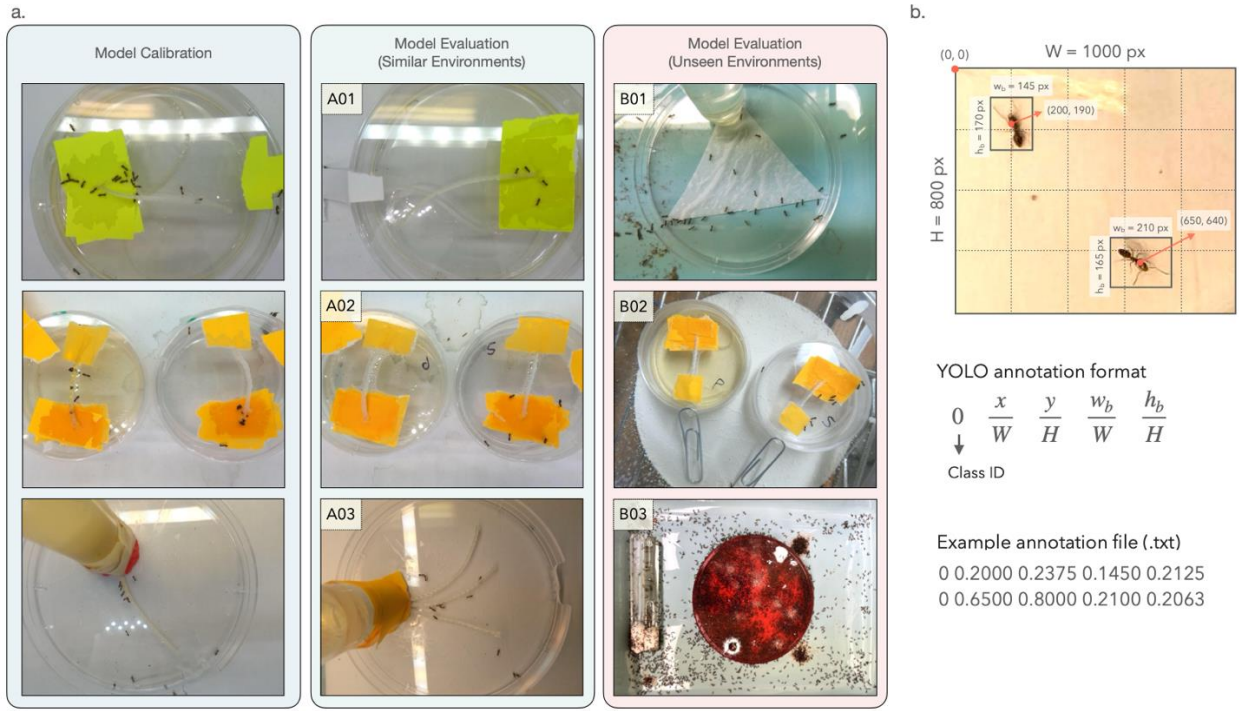

*Figure 1a.* Overview of the dataset subsets used in this study. *1b.* Example of ant annotation in the YOLO object detection format.

alt text: (a) Visual summary of the different data sets used in the study. (b) An image showing an ant marked with a box to demonstrate how the computer detects it.

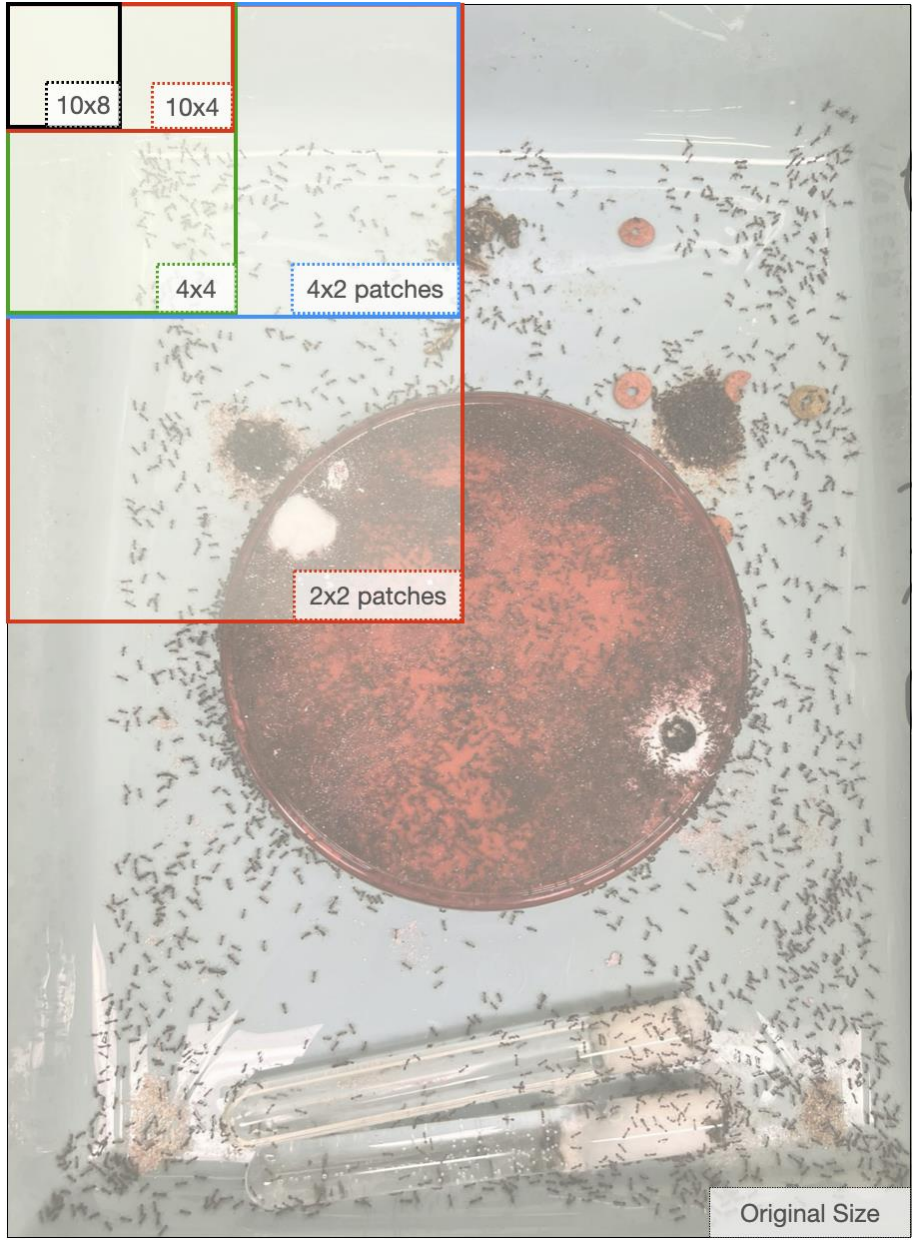

***Figure 2.*** *Illustration of the image slicing process for the "B03" subset. The original image is divided into patches of different sizes for object detection.*

alt txt: An illustration of how an original image from the "B03" set is divided into smaller pieces to help detect ants alt text: A graph showing how well the model performs on different data sets (indicated by different colors) using various numbers of calibration images. Shaded areas around each line show the variability in the results.

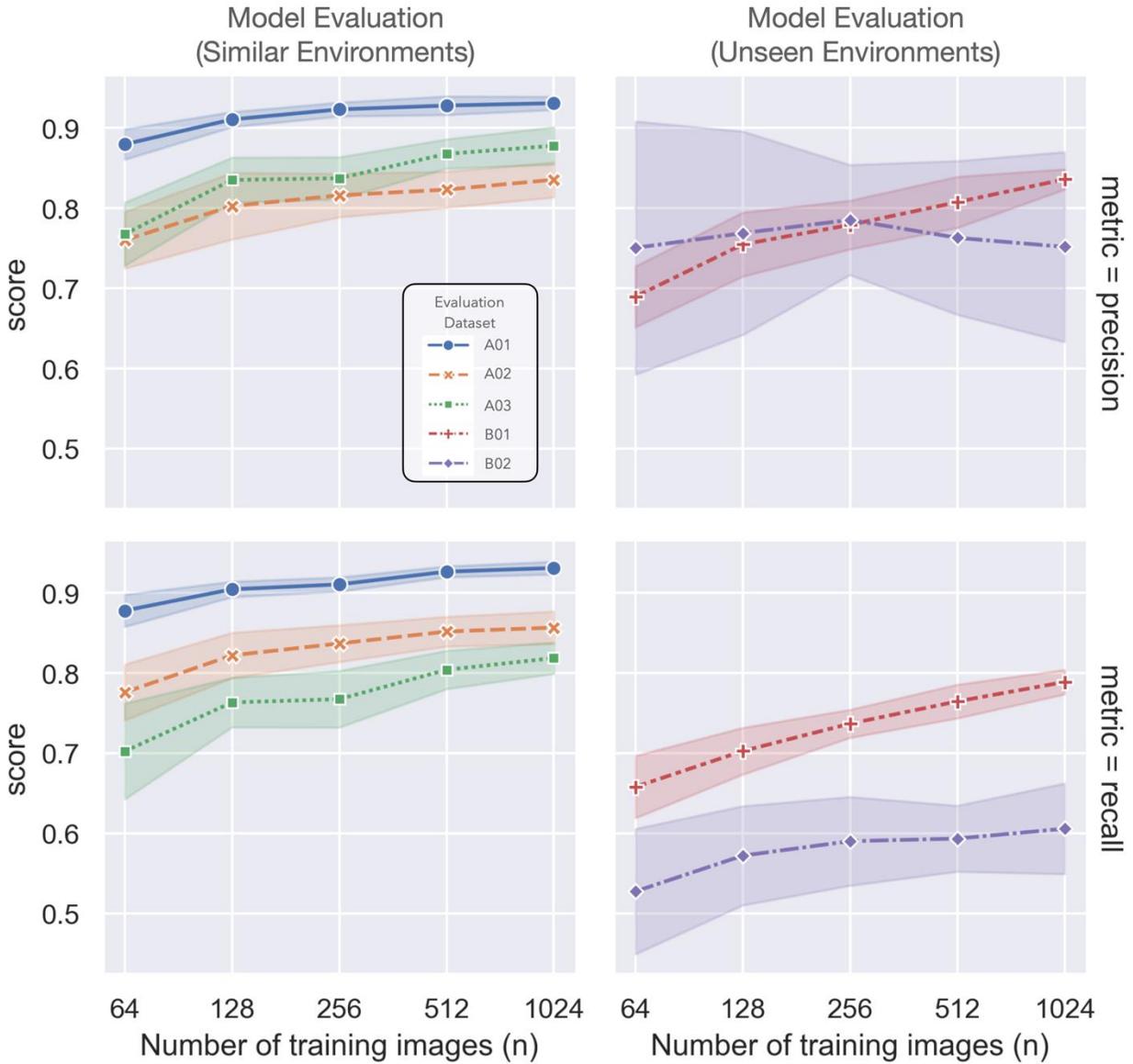

*Figure 3.* Model performance on different evaluation subsets (color-coded) with varying calibration set sizes (n). One standard deviation of the 30 sampling iterations is shown as colored bands of each line.

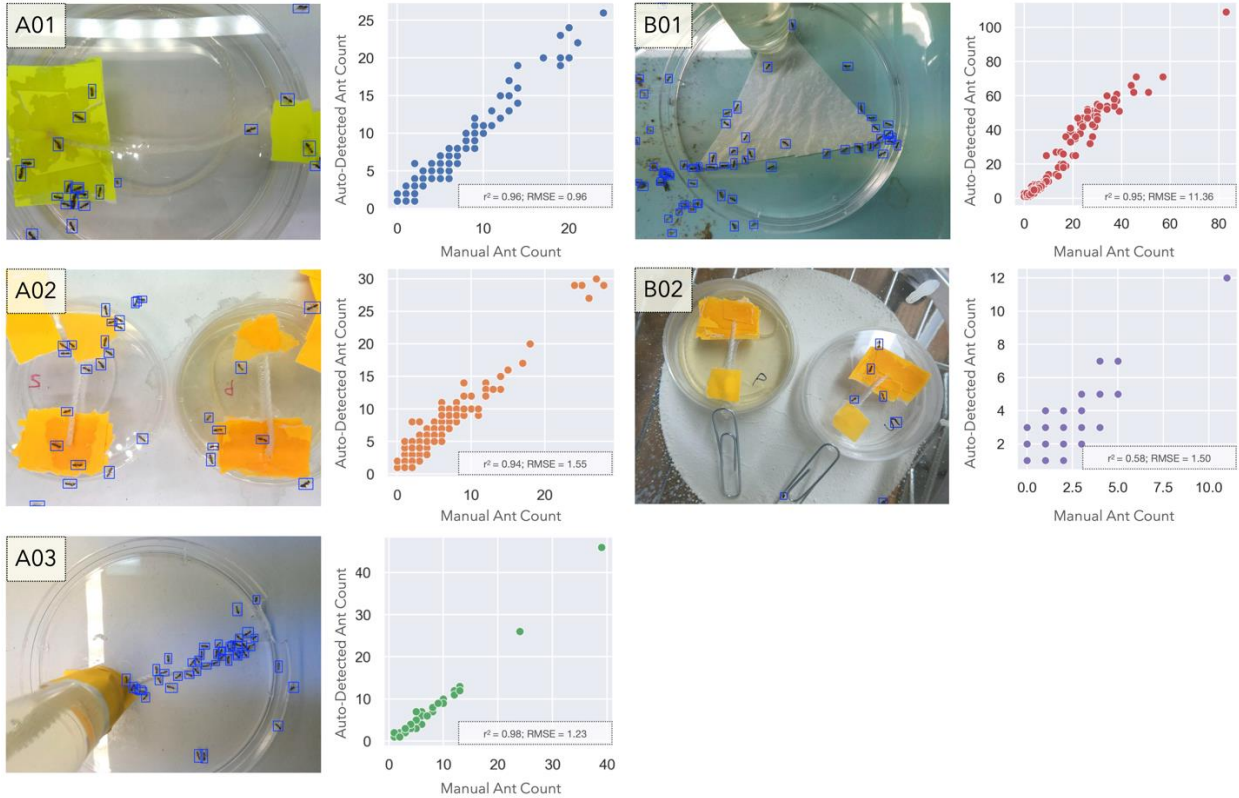

***Figure 4.*** *Comparison of manual and automated counting results for subsets A01, A02, A03, B01, and B02. The calibration set size is fixed at 1024 images. Each point represents the count of ants in a single image.*

alt text: A scatter plot comparing the number of ants counted manually versus counted by the computer for several data sets. Each dot represents one image's ant count.

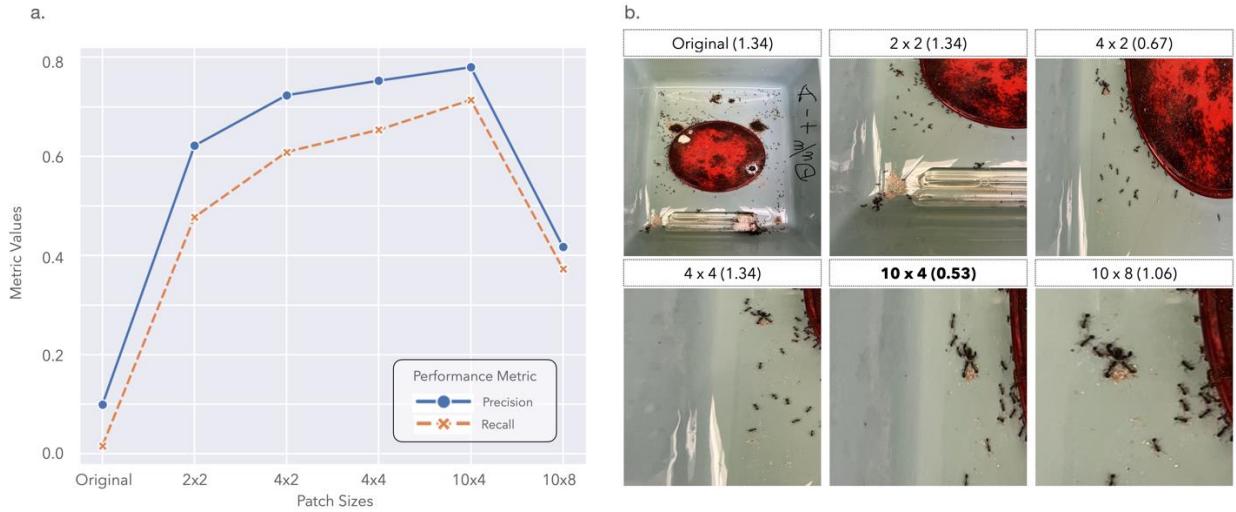

*Figure 5a.* Model performance on original and sliced images with different patch ratios. Precision and recall values are represented by blue and orange lines, respectively. *Figure 5b.* Example images resized to 640 x 640 pixels, with subtitles indicating the patch ratio and height-to-width ratio in parentheses.

alt text: 5a A line graph displaying the model's precision (blue line) and recall (orange line) when using original and sliced images with different patch sizes. 5b Examples of images resized to 640×640 pixels, each labeled with its patch size and aspect ratio.

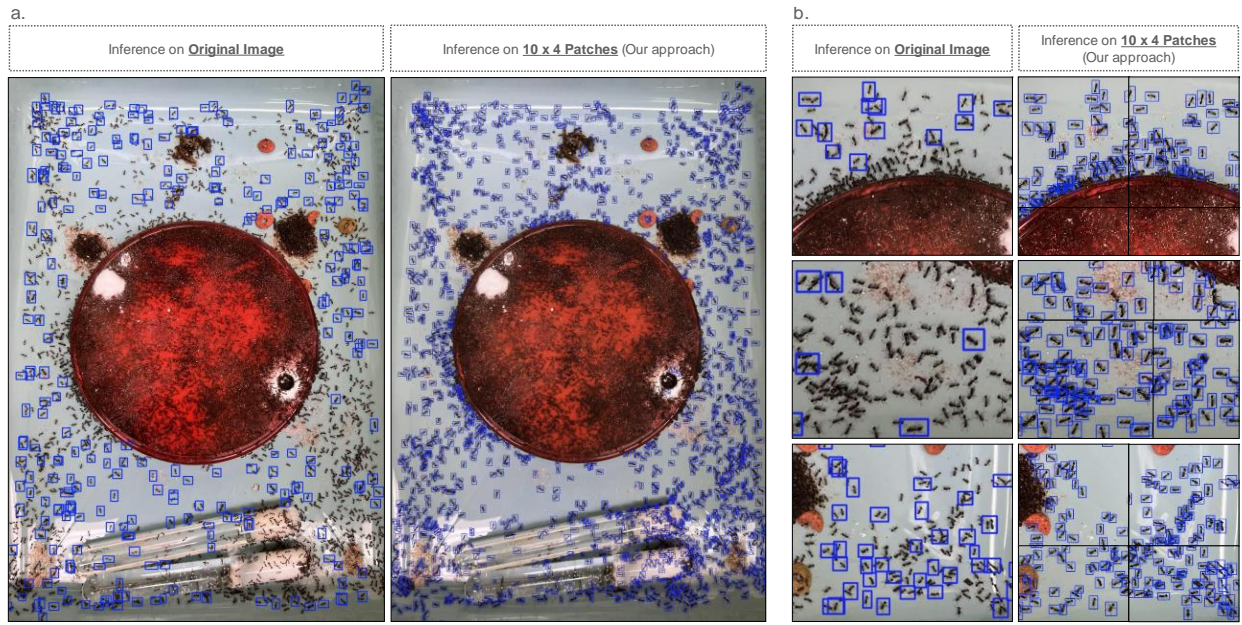

*Figure 6a.* Comparison of detection results between the original image and 10 x 4 patched images. The original image (left) and the 10 x 4 patched images (right) are displayed with blue detection bounding boxes. *Figure 6b.* Three examples in rows showing zoomed-in views of the original image (left) and the 10 x 4 patched images (right), each with blue detection bounding boxes.

alt text: 6a Side-by-side images showing ant detection results: the original image with blue boxes around ants on the left, and the same image divided into 10×4 patches with detections on the right. 6b Three sets of close-up views comparing the original image (left) and the patched images (right), both with ants highlighted by blue boxes.

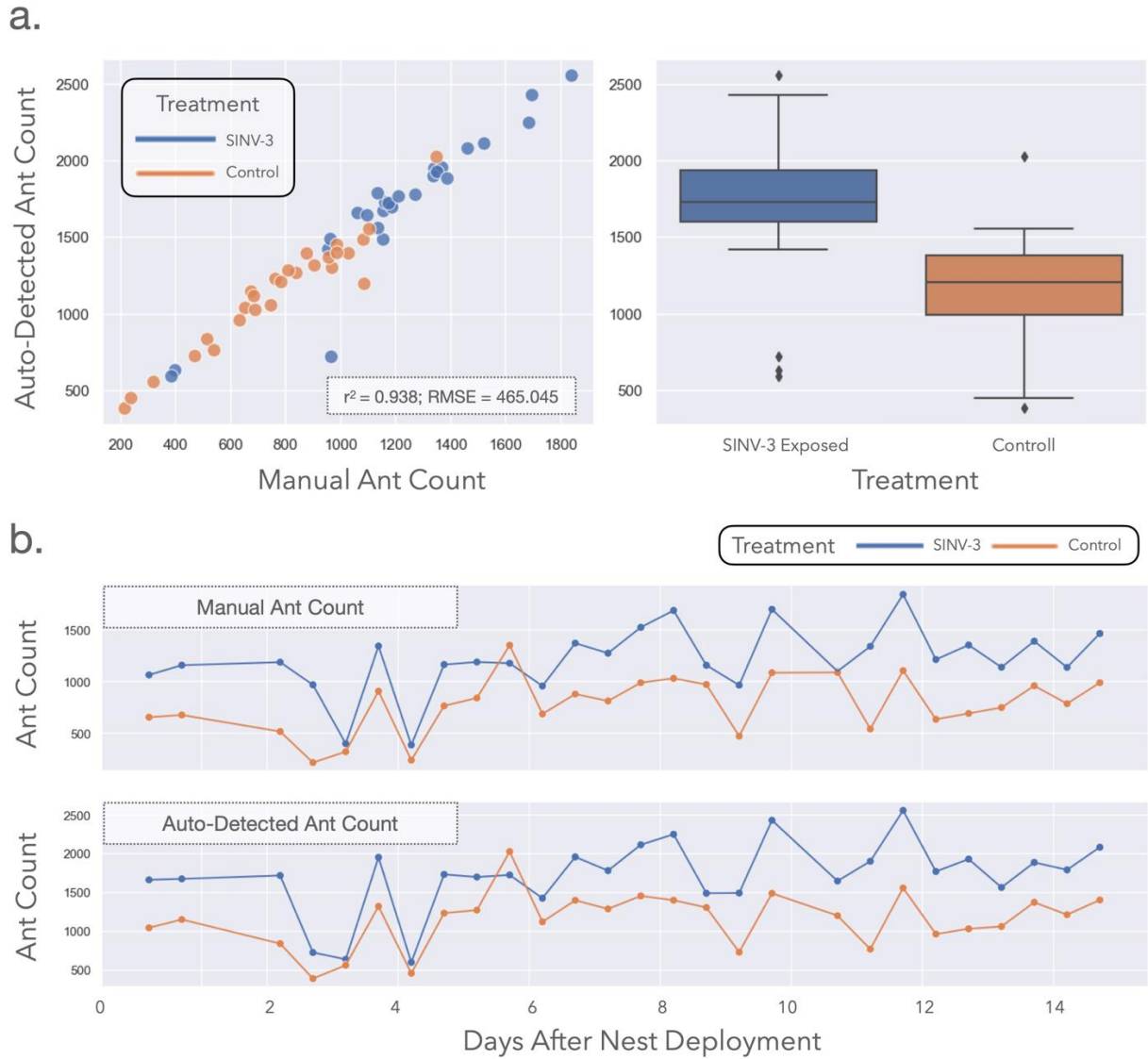

*Figure 7a.* Comparison of manual and automated counting results for the 10 x 4 patched images. The left panel shows a scatter plot of the manual and automated counts, while the right panel shows the automated counts for the SINV-3 infection experiment. *Figure 7b.* Temporal trend of ant population dynamics in the SINV-3 infection experiment.

alt text: 7a. Left: A scatter plot comparing manual and automated ant counts for the patched images. Right: A graph showing automated ant counts over time during the SINV-3 infection experiment. 7b. A line graph illustrating how the ant population changes over time in the SINV-3 infection experiment alt text: Visualizations of ant activity over time for experiments B01 (8a) and B02 (8b). Top row: Original images. Middle: Heatmaps showing where ants were active; brighter colors indicate more activity. Bottom row: Line graphs of ant counts over time.

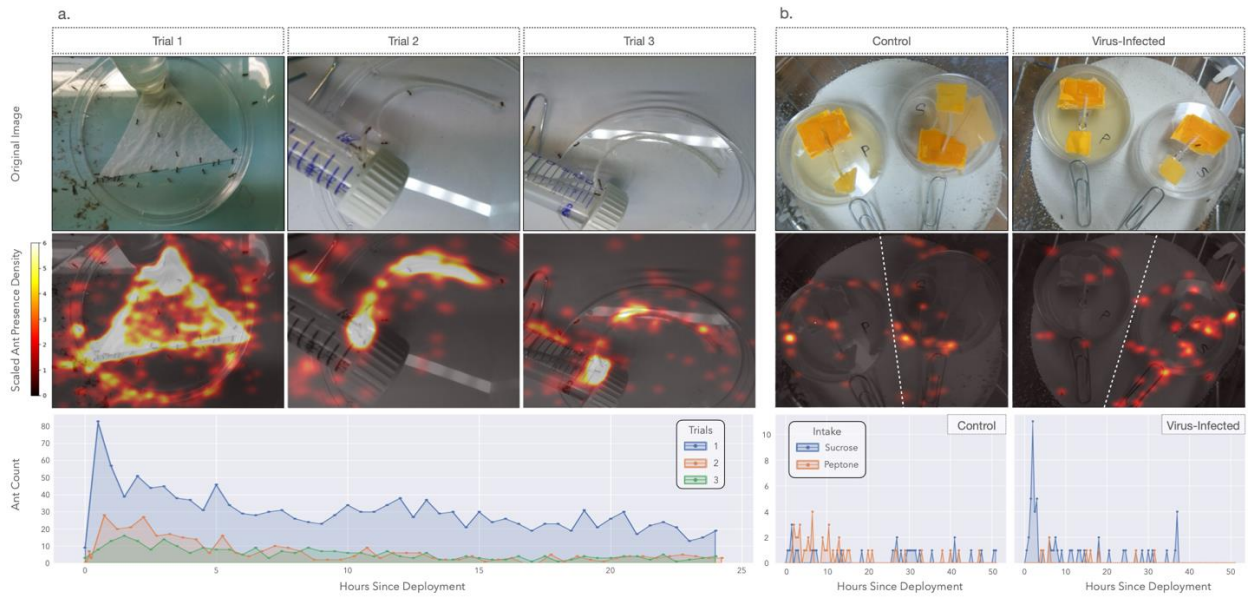

*Figure 8.* Heatmap of ant distribution over time for experiments B01 (8a) and B02 (8b). The top row displays the original images, followed by heatmaps illustrating ant activity over time. The color gradient represents the scaled number of ant detections, with white/yellow indicating high activity and black indicating low activity. The bottom row presents line plots of ant presence over time, with the x-axis representing hours and the y-axis representing the number of detected ants.

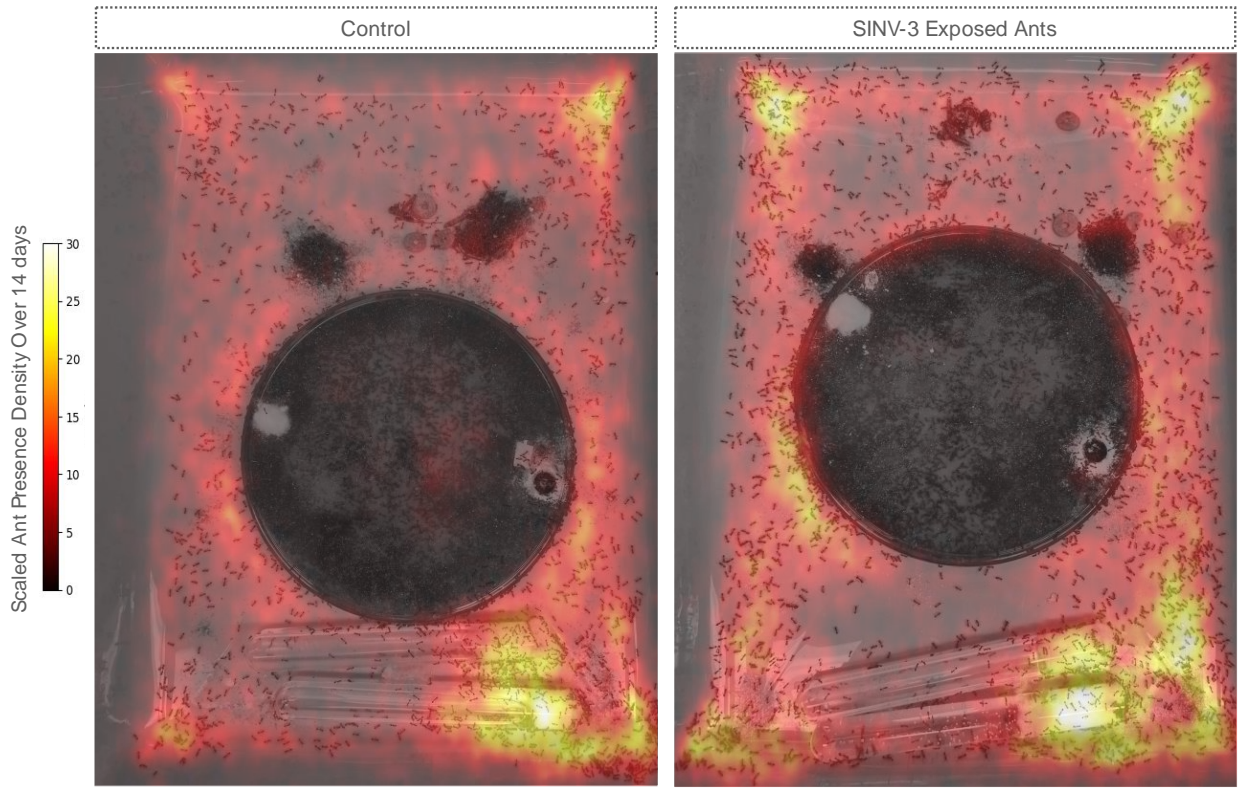

*Figure 9.* Heatmap of ant distribution in experiment B03. The left panel displays the *uninfected* group, while the right panel shows the SINV-3 infected group. The color gradient represents the scaled number of ant detections, with white/yellow indicating high activity and black indicating low activity.

alt text: Heatmaps comparing ant activity in experiment B03 between the uninfected group (left) and the SINV-3 infected group (right); brighter colors indicate higher ant activity levels.